\documentclass[conference]{IEEEtran}
\IEEEoverridecommandlockouts
\usepackage{cite}
\usepackage{amsmath,amssymb,amsfonts}
\usepackage{algorithmic}
\usepackage{enumitem}
\usepackage{graphicx}
\usepackage{textcomp}
\usepackage{xcolor}
\usepackage{bm}
\usepackage[T1]{fontenc}
\usepackage[utf8]{inputenc} % if your document is encoded in UTF-8
\usepackage[hyphens]{url}
\def\BibTeX{{\rm B\kern-.05em{\sc i\kern-.025em b}\kern-.08em
    T\kern-.1667em\lower.7ex\hbox{E}\kern-.125emX}}
\begin{document}

\title{VAAD: Visual Attention Analysis Dashboard applied to e-Learning
}

\author{\IEEEauthorblockN{
Miriam Navarro,
Álvaro Becerra,
Roberto Daza,
Ruth Cobos,
Aythami Morales,
Julian Fierrez}

\IEEEauthorblockA{School of Engineering, Universidad Autonoma de Madrid, Spain\\
\{miriam.navarroclemente,
alvaro.becerra,
roberto.daza,
ruth.cobos,
aythami.morales,
julian.fierrez\}@uam.es}
}

\IEEEoverridecommandlockouts
\IEEEpubid{\makebox[\columnwidth]{979-8-3503-7661-6/24/\$31.00~\copyright2024 IEEE \hfill} \hspace{\columnsep}\makebox[\columnwidth]{ }}

\maketitle
\IEEEpubidadjcol

\begin{abstract}
In this paper, we present an approach in the Multimodal Learning Analytics field. Within this approach, we have developed a tool to visualize and analyze eye movement data collected during learning sessions in online courses. The tool is named VAAD —an acronym for Visual Attention Analysis Dashboard—. These eye movement data have been gathered using an eye-tracker and subsequently processed and visualized for interpretation. The purpose of the tool is to conduct a descriptive analysis of the data by facilitating its visualization, enabling the identification of differences and learning patterns among various learner populations. Additionally, it integrates a predictive module capable of anticipating learner activities during a learning session. Consequently, VAAD holds the potential to offer valuable insights into online learning behaviors from both descriptive and predictive perspectives.

\end{abstract}


\section{Introduction}
The progression of technology has played a crucial role in fostering innovative education within online settings, alongside various other noteworthy advancements. A notable instance of this is the growing prominence of Massive Open Online Courses (MOOCs). Known for their diversity, MOOCs prioritize inclusivity and accessibility, as they do not necessitate specific prerequisites for enrollment. This wide array of subject matter makes them a valuable source of knowledge, as they are recognized by official educational institutions \cite{garcia2017mooc, ma2019investigating}. 

Educators can now upload learning materials, such as videos and other educational resources, through these online courses, making it easy for learners to access this information. This enables learners from all over the world to access numerous courses uploaded by different universities globally.

While the popularity of MOOCs continues to grow, so do dropout rates. Despite being an appealing learning tool, few learners successfully complete these courses \cite{cobos2017predicting}. This is why educators are worried about this trend, emphasizing the need to understand the underlying causes. 

This concern has led to the creation of Learning Analytics (LA) tools, which provide valuable insights into online educational environments \cite{lang2017handbook}. It is the case of Universidad Autónoma de Madrid (UAM), with the development of such systems, like edX-LIMS \cite{cobos2023self, cobos2021improving, cobos2020proposal, pascual2022proposal} or M2LADS \cite{becerra2023estudio, becerra2023m2lads}. 

The first one, edX-LIMS, gives learners feedback on their interaction with the MOOC, assisting them throughout their learning journey and addressing their problems, thereby keeping them engaged. They also provide instructors with a web-based Dashboard to monitor their learners, giving them a better understanding of their progress and helping them identify those in need of additional support. Consequently, instructors can intervene and offer targeted assistance to learners who may benefit from it.

The second one, M2LADS, facilitates the integration and visualization of multimodal data captured during monitoring using biometric sensors, presented through Web-based Dashboards.

With the main aim to complement and provide analysis of data managed by e-learning platforms, we have developed a tool designed to visualize eye movement data collected. The name of this tool is VAAD, which stands for Visual Attention Analysis Dashboard. This tool presents visualizations through different graphs such as boxplots and heatmaps to help understand the monitored individuals' behavior. In this article, we use data gathered during a learning session (LS) in online learning to better understand learners' behavior. However, VAAD could also be applied in other contexts involving web platforms and services, such as marketing studies or system analysis \cite{bialowkas2019eye}, to gain insights into users' preferences, or in gaming to improve players' performance.

The capture of data throughout learning sessions is carried out via the edBB platform, tailored for remote education, which captures both biometric and behavioral data \cite{hernandez2019edbb, daza2023edbb}.

In this article we present the tool VAAD applied to e-learning. The structure of this article is as follows: In the following section we present related works and the motivation for the approach proposed. Then, in Section \ref{sec3} we provide a description of the context and dataset. In Section \ref{sec4}, a detailed explanation of the proposed approach is presented. Finally, in Section \ref{sec5} the article concludes with conclusions and future work. 

\section{Related Works and motivation for the approach proposed}

Capturing and analyzing biometric multimodal data can be an effective method for gaining insight into a course's dynamics and learner engagement \cite{saqr2020multimodal}. For instance, audio data can be used to create a network of learner interactions; visual and eye-tracking data can offer insights into visual attention; sensors can measure psychological responses. If properly interpreted, these data can help understand interactions that contribute to educational success \cite{spikol2018supervised}.

Biometric multimodal data can be collected from various devices \cite{giannakos2022multimodal}, including webcams, electroencephalogram bands (EEG) for brain waves and attention levels, smartwatches for heart rate and EDA, or eye-trackers for pupil diameter and visual attention, including fixation gaze and saccades.

Regarding visual attention, earlier studies have shown that eye-tracking techniques can give a deeper understanding of learners' performance, regardless of their backgrounds \cite{sharma2016gaze}. Employing this approach could enhance attention and the quality of learning in MOOCs, due to the established correlation between learner performance and visual attention \cite{sharma2016visual}. 

In \cite{sharma2016gaze}, learners' visual attention was examined, and feedback on their visual performance led to a consistent improvement of their performance by 1\% every minute. 
Additionally, research in \cite{richardson2007art} revealed that listeners who focused more on visual references made by an interlocutor had higher comprehension levels, while research in \cite{mayer2010unique} found a strong correlation between fixation eye movements and learning outcomes.

To better understand visual data, visualization is crucial. In the LA field, the frequent use of dashboards \cite{verbert2013learning, verbert2014learning} aids learners and instructors in decision-making by providing a clear insight into the ongoing learning process \cite{martinez2012interactive}. 

Research cited in \cite{andreu2016ealab} highlights the effectiveness of dashboards in the aftermath of processing and synthesizing visual attention data, providing analysts with precise and succinct information. The study presents a dashboard designed to organize eye-tracking data from a range of learners, including information on eye fixations, saccades, and pupil diameter. Moreover, it successfully predicted with a 76.4\% accuracy whether a learner was fatigued or not employing predictive algorithms.

The eye-tracking technology is an integral part of all the research cited. This is why in the case of the tool VAAD  the focus remains on eye movement data.  This tool focuses on fixation and saccade events, as prior research has revealed that eye movements can indicate patterns and visual behaviors associated with different areas of interest (AOI) for learners, highlighting areas where they focus the most \cite{goodwin2022veta}. Moreover, this research has shown that studying saccade movements can successfully help identify AOIs.

VAAD's goal is to extract essential information about fixation and saccade events. This tool then presents visual data in two views: a global one, offering insight into the session and MOOC activities, and an individual view displaying learners' AOIs through a heat map.  Consequently, VAAD is equipped with a descriptive analysis module that includes the option for an ANOVA test, enabling it to identify correlations among different learner populations as seen in prior studies \cite{sharma2016application}. 

However, VAAD is not limited to a descriptive module; it also features a predictive analysis module that can identify what a learner is doing at a specific moment during the LS. This allows VAAD to provide insights using both descriptive and predictive approaches, enabling analysts to draw conclusions and gain a better understanding of the cognitive processes related to eye movement.

\section{Context and Dataset}
\label{sec3}
In order to test the approach proposed, we have conducted a study in the School of Engineering at our institution, where 120 learners from the school were monitored in our laboratory while they attended and interacted with a MOOC subunit during a 30-minute LS. The chosen MOOC was titled "Introduction to Development of Web Application" (WebApp), which is available on the edX MOOC platform\footnote{\url{https://www.edx.org/}} and offered by our university. 

In this study, the monitored learners engaged in various activities such as watching videos, reading documents, and then completing assignments on the LS content, which was about HTML language. Prior to the gathering of multimodal data from the LS, learners underwent a pretest to determine their initial level of knowledge on HTML. At the end of the LS, the assignments that had been completed by the learners in the MOOC served as the posttest items.

Learners were monitored throughout the LS via the edBB platform. All multimodal biometric data captured by their sensors are synchronised by M2LADS \cite{becerra2023estudio, becerra2023m2lads}. The sensors used to obtain the different multimodal data are as follows:
\begin{itemize}[label={\LARGE{\bm{$\cdot$}}}]
    \item Video data: Acquired from overhead, front and side cameras, consisting of 2 webcams and 1 Intel RealSense.  Additionally, video from the screen monitors was also recorded.
    \item Electroencephalogram (EEG) data: Acquired from a NeuroSky EGG band \cite{daza2022alebk,daza2023matt}.
    \item Heart rate data: Acquired from a Huawei Watch 2 pulsometer feature \cite{hernandez2020heart}.
    \item Visual attention data: Acquired from a Tobii Pro Fusion that contains 2 eye-tracking cameras. 
\end{itemize}

As previously mentioned, VAAD centers on managing all multimodal data linked to visual attention, which is why the Tobii Pro Fusion eye-tracking device remains the focal point. The data for VAAD arises from the processing module, which handles the eye-tracking data produced by the eye-tracker, as well as additional data generated by M2LADS. This study has received approval from the university's ethics committee, and all biometric multimodal data are anonymized.

\subsection{Eye-tracking Data}

The eye-tracker estimates a wide variety of multimodal data. However, for the development of VAAD, the processing module only works with specific metrics, which are listed in Table \ref{t1}.

\begin{table}[h]
\centering
\caption{Multimodal data extracted from the Tobii Pro Fusion eye-tracking device}
\begin{tabular}{|p{2cm}|p{3.5cm}|p{2cm}|}
    \hline
    \textbf{Variable} & \textbf{Description} & \textbf{Units}\\
    \hline
    Participant ID & Anonymized ID & -\\
    \hline
    Recording date & - & YYYY-MM-DD\\
    \hline
    Recording start time & - & HH:MM:SS:mmm\\
    \hline
    Recording timestamp & Timestamp counted from the start of the recording ($t_0=0$) & Milliseconds\\
    \hline
    Gaze point X & Horizontal coordinate of the averaged left and right eye gaze point & Pixels \\
    \hline
    Gaze point Y & Vertical coordinate of the averaged left and right eye gaze point & Pixels \\
    \hline
    Eye movement type & Type of eye movement & Fixation, Saccade, Unclassified, Eyes Not Found\\
    \hline
    Gaze Event duration & The duration of the currently active eye movement & Milliseconds\\
    \hline
    Eye movement type index & Count is an auto-increment number starting with 1 for each eye movement type & Number\\
    \hline
\end{tabular}
\label{t1}
\end{table}

\subsection{M2LADS Data}
Additional data regarding learner population and the LS are obtained from the M2LADS database. The processing module exclusively operates with specific data, which are listed in Table \ref{t2}.

Prior to the monitorization, learners were categorized into three different groups (40 learners per group), which determined the frequency of interruptions they encountered during the LS.

\begin{table}[h]
\centering
\caption{Data extracted from M2LADS}
\begin{tabular}{|p{3.5cm}|p{4.5cm}|}
    \hline
    \textbf{Variable} & \textbf{Description} \\
    \hline
    Participant ID & Anonymized ID of the participant\\
    \hline
    Activity identifier & Activity of the MOOC in which the learner is involved \\
    \hline
    Initial time & Initial timestamp of the activity \\
    \hline
    Final time & Final timestamp of the activity \\
    \hline
    Sex & Participant sex \\
    \hline
    Group & Participant group \\
    \hline
    HTML level & Initial knowledge level on HTML \\
    \hline
    Academic background & University degree\\
    \hline
    Learning score & Difference between the posttest and pretest grade \\
    \hline
\end{tabular}
\label{t2}
\end{table}

\section{Approach proposed}
\label{sec4}
\begin{figure*}[t]
 \centering
  \includegraphics[width=0.7\linewidth]{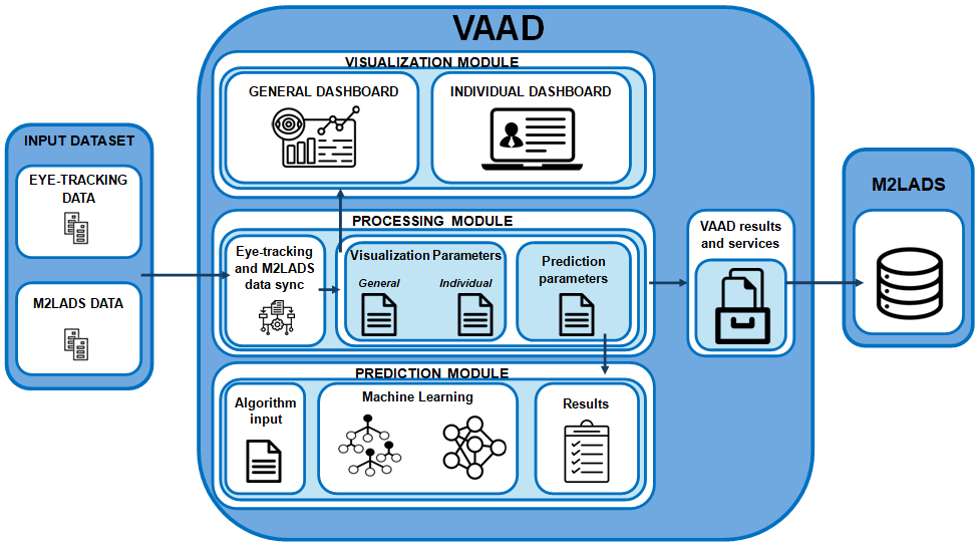} %./Figures/.png
  \caption{VAAD Architecture/Modules}
   \label{VAAD_arch}
\end{figure*}

The VAAD tool is composed of three different modules: a processing data module, a visualization module, and a prediction module. As seen in Fig. \ref{VAAD_arch}, the processing module manages the data from the eye-tracker and M2LADS, and is interconnected with both the visualization and prediction modules, which use the processed data for their respective tasks. Finally, all preprocessed data and results are stored in M2LADS database.

\begin{figure*}[t]
 \centering
  \includegraphics[width=\linewidth]{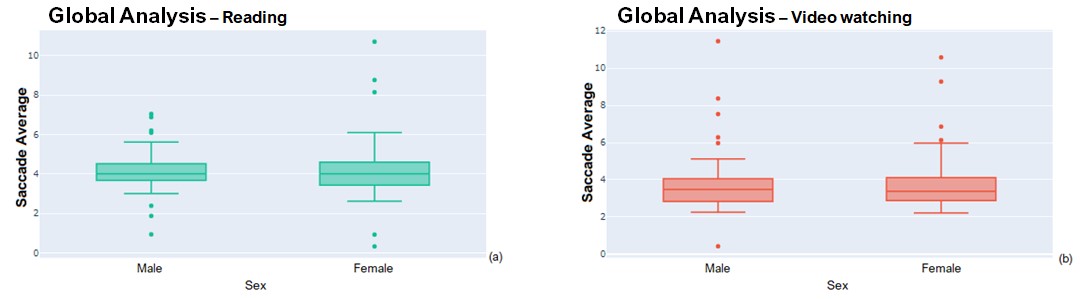} %./Figures/.png
  \caption{Screenshot from an example overview of global analysis for reading actuvity (a) and video watching (b)}
   \label{VIS2}
\end{figure*}

\begin{figure*}[t]
 \centering
  \includegraphics[width=\linewidth]{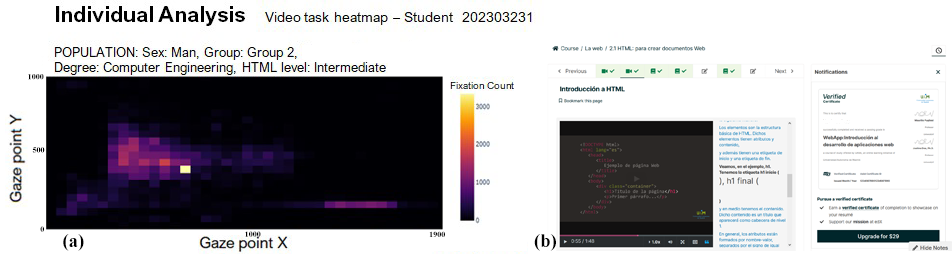} %./Figures/.png
  \caption{Screenshot from an example overview of individual analysis  (a) and screenshot from the video that the learner is watching (b)}
   \label{VIS}
\end{figure*}

\subsection{Processing Data Module}
Throughout the LS, VAAD has specifically focused on capturing saccade and fixation event information provided by the eye-tracker, in addition to timestamps and event durations, to gather insights into these eye movements.

Before the processing stage, the eye-tracking data of learners were thoroughly inspected to identify errors or anomalies. Learners displaying a notable frequency of "Eyes Not Found" values for the Eye movement type variable were excluded to ensure the integrity of the final data utilized by VAAD.

The objective of the processing module was to synchronize the eye-tracking data with the M2LADS data, ensuring that each learner's eye-tracking data were accurately tagged with the corresponding activities within the MOOC for every frame captured. This approach enabled us to determine precisely when each eye event occurred during the LS, utilizing the available timestamps and activities. By incorporating this tagging process, we were able to generate the necessary data for the visualization module and the prediction module.

From a general visualization perspective, data were compiled to include all learners, incorporating their sex and group, thus enabling differentiation of populations based on these factors. Subsequently, four key parameters were determined for each learner: average saccades, average fixations, average saccade time, and average fixation time. These parameters were not only assessed for the overall session but also for each activity individually. The culmination of this data processing is a final database that profiles each learner, including their respective group and sex, and provides details on the four aforementioned parameters for every activity within the MOOC, as well as for the entire session.

Regarding the individual visual overview, data were generated for each available learner, which included information about their saccade and fixation events, along with their gaze point parameters for the X and Y coordinates and the activity within the MOOC. 

Concerning the predictive analysis, data were utilized to collect various metrics subsequently employed for prediction. Detailed information on these metrics is gathered in Table \ref{t3}.

\begin{table}[h]
\centering
\caption{Metrics used in the prediction module}
\begin{tabular}{|p{3.5cm}|p{4.5cm}|}
    \hline
    \textbf{Feature} & \textbf{Description} \\
    \hline
    Sex & Participant sex\\
    \hline
    Label & The activity to be predicted within the MOOC \\
    \hline
    Saccade number & Average number of saccade movements \\
    \hline
    Velocity & Average saccade velocity \\
    \hline
    Velocity X & Average horizontal saccade velocity \\
    \hline
    Velocity Y & Average vertical saccade velocity \\
    \hline
    Max velocity & Maximum saccade velocity \\
    \hline
    Min velocity & Minimum saccade velocity \\
    \hline
    Deviation & Saccade velocity typical deviation \\
    \hline
    Deviation X & Saccade horizontal velocity standard deviation \\
    \hline
    Deviation Y & Saccade vertical velocity typical deviation \\
    \hline
    Kurtosis & Saccade velocity kurtosis \\
    \hline
    Kurtosis X & Saccade horizontal velocity kurtosis \\
    \hline
    Kurtosis Y & Saccade vertical velocity kurtosis \\
    \hline
    Skew & Saccade velocity skew\\
    \hline
    Skew X & Saccade horizontal velocity skew\\
    \hline
    Skew Y & Saccade vertical velocity skew\\
    \hline
\end{tabular}
\label{t3}
\end{table}

\subsection{Visualization Module}
The visualization charts crafted by VAAD provide a thorough examination of the session, providing a deep understanding of both global trends and individual learners' engagement. These charts are available in both English and Spanish.

The general session overview is presented through interactive box plot charts, providing a visual representation of the four parameters mentioned earlier, namely average saccades, average saccade time, average fixations, and average fixation time. These charts can be filtered by different demographic categories, such as groups or sex, allowing for a detailed exploration of the data. Fig. \ref{VIS2} (a) illustrates an example of average saccade movements during reading, filtered by sex, revealing a greater distribution among females. The same pattern is observed for video watching in Fig. \ref{VIS2} (b). Analysts have the option to visualize specific activities within the MOOC or view the entire session at once, with each task differentiated by a different color for enhanced clarity. Moreover, an ANOVA test is conducted for each of the four parameters, identifying significant variations among learners from different populations.

Another interactive chart presents individual learners' data through a heat map, providing a visual representation of their on-screen attention. As depicted in the example shown in Fig. \ref{VIS} (a), it is possible to identify precisely where the learner was looking during the video task, with a higher concentration of fixations observed within the region of the screen where the video was displayed (Fig. \ref{VIS} (b)). Analysts can select a specific learner and customize their visual screen heat map by choosing from various options, such as different activities within the MOOC session. This approach offers valuable insights into each learner's focus during different activities, assisting in the identification of engaging materials and assessing the impact of diverse learning sources on learners' performance.

\subsection{Prediction Module}
The goal of the prediction module is to determine the activity which the learner is doing. The identifier metric may indicate either video watching or course content reading. The prediction module utilizes the metrics outlined in Table \ref{t3} to make a binary prediction between reading and video watching.

For the prediction task, we worked with learners belonging to groups 2 and 3 (80 learners), excluding those from group 1 due to the high frequency of interruptions experienced by learners in this group during the LS. Additionally, we excluded some data from the reading category to balance our training dataset, ensuring an equal number of data points for each category (524 samples for each category). 

We used the Random Forest algorithm and a neural network to test against the testing data. The neural network tested is a perceptron with one hidden layer (32 neurons and ReLU activation) and one output layer (sigmoid activation). The loss function chosen was Mean Squared Error (MSE) with Adam optimizer (default learning rate of 0.001).

We adopted two distinct approaches. Initially, we divided the learners from groups 2 and 3 into 75\% for training data and 25\% for testing data. Subsequently, we used both Random Forest and the neural network for analysis. (See Table \ref{t5})

\begin{table}[htbp]
\centering
\caption{Results of testing two algorithms on the testing data with  75\% training and 25\% test}
\begin{tabular}{|c|c|c|c|}
    \hline
     \multicolumn{2}{|c|}{} & \textbf{Random Forest} & \textbf{Neural Network} \\
    \hline
     \multicolumn{2}{|c|}{Accuracy test} & 0.72 & 0.73\\
    \hline
    & Precision & 0.62 & 0.67\\
    \cline{2-4}
   Video watching & Recall & 0.88 & 0.75 \\
    \cline{2-4}
    & F1-Score & 0.73 & 0.70\\
    \hline
    & Precision & 0.88 & 0.80\\
    \cline{2-4}
   Reading & Recall & 0.61 & 0.74 \\
   \cline{2-4}
    & F1-Score & 0.72 & 0.77\\
    \hline
\end{tabular}
\label{t5}
\end{table}

The second approach consisted of evaluating both methods using the Leave-One-Out Cross-Validation (LOOCV) technique.

Following the testing of both approaches for both predictive methods, the results obtained after applying the LOOCV technique were more promising as we don´t have a lot of data. The results obtained from the LOOCV technique are presented in Table \ref{t4}.

\begin{table}[htbp]
\centering
\caption{Results of testing two algorithms on the testing data with LOOCV}
\begin{tabular}{|c|c|c|c|}
    \hline
     \multicolumn{2}{|c|}{} & \textbf{Random Forest} & \textbf{Neural Network} \\
    \hline
     \multicolumn{2}{|c|}{Accuracy test} & 0.76 & 0.74\\
    \hline
    & Precision & 0.76 & 0.74\\
    \cline{2-4}
   Video watching & Recall & 0.79 & 0.76 \\
    \cline{2-4}
    & F1-Score & 0.77 & 0.75\\
    \hline
    & Precision & 0.78 & 0.75\\
    \cline{2-4}
   Reading & Recall & 0.75 & 0.73 \\
   \cline{2-4}
    & F1-Score & 0.76 & 0.74\\
    \hline
\end{tabular}
\label{t4}
\end{table}

For LOOCV, in terms of accuracy, both methods show close similarity, with Random Forest slightly edging ahead. This trend persists across the other metrics, with Random Forest consistently showing a slight advantage.

Overall, while both models share similarities, Random Forest shows a slight superiority.

\section{Conclusions and Future work}
\label{sec5}
In this paper, we introduce VAAD (Visual Attention Analysis Dashboard), an innovative tool designed to visualize biometric data related to visual attention gathered from monitored sessions. In this article, we use eye movement data from a learning session and this tool enables analysts to gain a deeper understanding of learner behavior by filtering and visualizing different components of the session. 

It offers valuable insights into learners' focus and engagement through the analysis of eye movements. Moreover, it provides the flexibility to filter data by various learner demographics, enabling a more detailed exploration of the data. Additionally, the tool facilitates ANOVA tests, allowing for the identification of significant differences among learners during the learning process. The data managed by VAAD also offer the opportunity to detect the tasks performed by online learners during the learning session, which can be valuable information for instructors. 

This tool has the potential to significantly enhance the analysis of online learning behaviors and provide valuable insights for educational practitioners, and is currently being used by the MOOC instructors' team to gather information on learner visual behaviour while learning.

Finally, future work will further explore current and emerging predictive methods to ascertain the most suitable model for task prediction and investigate other indicators that have proven to be useful in predicting activities such as eyeblink \cite{daza2020mebal, daza2024mebal2}, eye pupil size \cite{rafiqi2015pupilware}, keystroking \cite{morales2016kboc}, among others. Additionally, we will tested the use of this tool in others contexts where the behavior of users looking at the screen is to be analyzed.

\section*{Acknowledgment}
Support by projects: HumanCAIC (TED2021-131787B-I00 MICINN), SNOLA (RED2022-134284-T), e-Madrid-CM (S2018/TCS-4307, a project which is co-funded by the European Structural Funds, FSE and FEDER), IndiGo! (PID2019-105951RB-I00), TEA360 (PID2023-150488OB-I00, SPID202300X150488IV0), BIO-PROCTORING (GNOSS Program, Agreement Ministerio de Defensa-UAM-FUAM dated 29-03-2022), and Cátedra ENIA UAM-VERIDAS en IA Responsable (NextGenerationEU PRTR TSI-100927-2023-2). Roberto Daza is supported by a FPI fellowship from MINECO/FEDER.

\bibliographystyle{IEEEtran}
\bibliography{bibliography}

\end{document}